\documentclass[runningheads]{llncs}
\usepackage{graphicx}
\usepackage{hyperref}
\usepackage{amsfonts}
\usepackage{amsmath}
\usepackage{bm}
\usepackage{booktabs}
\usepackage{subfig}
\usepackage{tabularray}
\UseTblrLibrary{booktabs}

\let\oldeqref\eqref
\renewcommand{\eqref}[1]{Eq.~\oldeqref{eq:#1}}

\newcommand\secref[1]{Section~\ref{sec:#1}}
\newcommand\figref[1]{Fig.~\ref{fig:#1}}
\newcommand\tabref[1]{Table~\ref{table:#1}}

\newcommand\LtoR[1]{\overrightarrow{#1}}
\newcommand\RtoL[1]{\overleftarrow{#1}}

\newcommand\CM{\checkmark}
\newcommand\trans[1]{#1^\top}
\newcommand\SoftMax[1]{\mathrm{SoftMax}(#1)}
\newcommand\LayerNorm{\mathrm{LayerNorm}}
\newcommand\EditDist{\mathrm{EditDist}}

\newcommand\SOS{\texttt{SOS}}
\newcommand\EOS{\texttt{EOS}}
\newcommand\SEP{\texttt{SEP}}
\newcommand\PAD{\texttt{PAD}}

\newcommand\B{\texttt{<b>}}
\newcommand\TD{\texttt{<td>}}
\newcommand\TDE{\texttt{</td>}}
\newcommand\TDL{\texttt{<td}}
\newcommand\colspan{\texttt{colspan="2"}}
\newcommand\rowspan{\texttt{rowspan="3"}}
\newcommand\TDR{\texttt{>}}

\pdfpagesattr{/CropBox [86 65 526 731]}

\gdef\papertitle{Multi-Cell Decoder and Mutual Learning for Table Structure and Character Recognition}
\title{\papertitle}
\titlerunning{Multi-Cell Decoder and Mutual Learning for Table Recognition}

\author{Takaya Kawakatsu\inst{1}}
\institute{
Preferred Networks, Inc., 1-6-1 Otemachi, Chiyoda, Tokyo, Japan.\\
\email{kat.nii.ac.jp@gmail.com}\\
\url{https://researchmap.jp/t.kat}
}

\hypersetup{pdftitle={\papertitle},pdfauthor={Takaya Kawakatsu},hidelinks}

\begin{document}
\maketitle

\begin{abstract}
Extracting table contents from documents such as scientific papers and financial reports and converting them into a format that can be processed by large language models is an important task in knowledge information processing.
End-to-end approaches, which recognize not only table structure but also cell contents, achieved performance comparable to state-of-the-art models using external character recognition systems, and have potential for further improvements.
In addition, these models can now recognize long tables with hundreds of cells by introducing local attention.
However, the models recognize table structure in one direction from the header to the footer, and cell content recognition is performed independently for each cell, so there is no opportunity to retrieve useful information from the neighbor cells.
In this paper, we propose a multi-cell content decoder and bidirectional mutual learning mechanism to improve the end-to-end approach.
The effectiveness is demonstrated on two large datasets, and the experimental results show comparable performance to state-of-the-art models, even for long tables with large numbers of cells.
\end{abstract}

\section{Introduction}

Information retrieval technology which provides high-quality knowledge to large language models (LLMs) is attracting attention.
Many researchers have worked on converting scanned and imaged documents to machine-readable formats such as HTML code~\cite{ICDAR21,Zhong20,Li20} and LaTeX code~\cite{Deng19,LaTeX21}.
This initiative has direct and indirect benefits.
First, because past literature remains mostly in printed form, it is necessary to convert it into structured electronic documents.
This is a direct benefit.
Second, the realization of intelligence which recognizes hidden meanings in the layout of published documents that are intended to be read by humans is important from the perspective of human-machine interaction.
This is an indirect benefit.

In this paper, we will focus on table recognition, which includes two types of tasks, namely structure recognition and cell content recognition.
A simple table has horizontal and vertical borders, and each cell contains characters.
Complex tables may contain cells which are merged vertically or horizontally, and/or cells which involve invisible borders.
Persons can understand table structure from the cell layout even without explicit boundaries, and this is a challenging problem.

In recent years, following the success of Transformer~\cite{Tr17} models in language and visual recognition tasks, many methods~\cite{Nassar22,VCG21,Nam23LA,Nam23GA} based on Transformer have been proposed for the table recognition task.
Since we can utilize an external optical character recognition (OCR) system to parse cell contents, we can focus mainly on table structure recognition.
This task exploits cross attention between image features and embedded representations of HTML tokens to predict HTML tokens sequentially.
In previous studies~\cite{Nassar22,VCG21,Nam23LA,Nam23GA,Nam23WS}, the token prediction was performed in one direction from the header to the footer, from left to right.
This impairs the opportunity to focus on the table structure ahead.

Ly and Takasu~\cite{Nam23GA} reported that end-to-end learning of table structure and cell content recognition tasks may improve overall table recognition performance.
In addition, tables may contain more than a few hundred cells, and the sequential prediction approaches may suffer from poor performance, which can be improved by introducing local attention~\cite{Nam23LA}.
These previous studies performed cell content recognition independently for each detected cell after structure recognition.
This impairs the opportunity of obtaining useful information from neighbor cells.

As a solution to the problems, we improve the end-to-end approach~\cite{Nam23LA,Nam23GA,Nam23WS} and propose a method that refers to the recognition results of neighbor cells and a learning mechanism focusing on both previous and following cells.
The former is achieved by introducing a cell decoder that infers multiple cells and configuring a hierarchical decoder along with an HTML decoder for structural recognition.
The latter is achieved by mutual learning~\cite{Mut18} between a forward decoder which reads the table structure from left to right and a backward decoder which reads the table structure from right to left.
The effectiveness of the proposed method is demonstrated using two large-scale tabular image datasets.

The main contributions of this paper are:
1) We propose a cell decoder which infers multiple cells and obtains useful information from surrounding cells.
2) We propose a bidirectional mutual learning mechanism to force the proposed model to pay attention to both previous and following cells.
3) Across all experimental results, our proposed method achieved better performance than state-of-the-art models.

\section{Related Work}

In general, the table recognition task is performed by two subtasks, namely table structure recognition and cell content recognition.
Of course, the final output is an HTML~\cite{ICDAR21,Zhong20,Li20} or LaTeX~\cite{Deng19,LaTeX21} document, so there is no need to distinguish between the two subtasks.
However, it is better to recognize tags (or commands) and other visible characters using separate models.
For cell content recognition task, existing highly accurate OCR systems~\cite{Lu21} are available, and thus previous studies~\cite{Ito93,Kien98,Wang04,Pra20,Raja20,Sch17} have mainly focused on table structure recognition.

Table structure recognition has been studied for a long time, and approaches based on hand-crafted features and heuristic rules~\cite{Ito93,Kien98,Wang04} were proposed, but their application was limited to simple tables or tables with predefined patterns.
With the development of deep learning, methods that automatically learn table structural patterns~\cite{Zhang22,Qiao21,Pra20,Raja20,Sch17} have become mainstream.
These studies can be divided into approaches based on object detection and segmentation~\cite{Sch17,Qiao21}, and approaches based on sequential token prediction~\cite{Nassar22,VCG21}.

For the detection and segmentation approaches, Schreiber et al.~\cite{Sch17} proposed a two-fold system using Faster R-CNN~\cite{RCNN15} and fully convolutional networks~\cite{FCN15} for both table detection and table structure recognition.
Raja et al.~\cite{Raja20} proposed a two-stage model that estimates the relationships between cells after recognizing their locations.
Qiao et al.~\cite{Qiao21} won his first place in the ICDAR competition~\cite{ICDAR21} by combining text, cell, row and column recognition tasks using Mask R-CNN~\cite{RCNN17}.

For the sequential token prediction approaches, a simple image caption model can be utilized for cell detection because the order of cells is uniquely determined.
Ye et al.~\cite{VCG21} and Nassar et al.~\cite{Nassar22} proposed Transformer models with two types of decoders for table structure recognition and cell localization.
Peng et al.~\cite{Peng23} achieved performance comparable to a model using a deep convolutional encoder while significantly reducing parameters by introducing a convolutional stem.

In the 2020s, researchers are investigating end-to-end models that learn both table structure and cell content recognition tasks~\cite{Deng19}.
Zhong et al.~\cite{Zhong20} proposed a model that uses a ResNet~\cite{Res15} encoder and two LSTM~\cite{LSTM97} decoders to recognize both table structure and cell contents, but its performance was inferior to models using external OCR.

Ly and Takasu~\cite{Nam23GA} proposed a multi-task model that detects table structure, cell locations, and cell contents.
Their model uses a ResNet encoder with global context attention~\cite{GCA19} and two Transformer decoders.
The first decoder infers the HTML tokens sequentially, and then the second decoder reads the cell contents one by one.
This model achieved performance comparable to the models utilizing external OCR.
They also proposed weakly supervised learning to reduce the cost of preparing bounding box training data~\cite{Nam23WS} and introduced local attention~\cite{Tr20} to effectively recognize tables with a large number of cells~\cite{Nam23LA}.

In 2021, the scientific literature parsing competition~\cite{ICDAR21} was held at ICDAR 2021.
The competition consisted of document layout recognition task A and table recognition task B.
Task B required converting table images to HTML tags with cell contents.
The PubTabNet~\cite{Zhong20} dataset and the final evaluation dataset were provided for this task.
The training dataset consists of HTML tokens, cell texts, and cell bounding boxes.
There were 30 submissions from 30 teams and most of the top 10 solutions exploited separate OCR models, additional annotation and ensemble techniques.

TabRecSet~\cite{Yang23} is a bilingual dataset containing rotated and distorted tables in real photographs for three tasks, namely table detection, structure recognition, and cell content recognition.
Detection of such tables is outside the scope of this paper.

\section{Background}

Similar to previous work~\cite{Nam23LA,Nam23GA,Nam23WS}, our proposed model uses a ResNet encoder and an HTML decoder consisting of multiple attention blocks~\cite{Tr17} to infer HTML tokens representing table structure.
An additional decoder is exploited to infer cell contents.
The encoder and two decoders are trained simultaneously using an end-to-end approach.
In this section, we introduce some techniques used by the proposed method described in \secref{mutab}.

\subsection{Encoder}

Previous studies~\cite{Nassar22,VCG21,Nam23LA} used a convolutional neural network (CNN) to extract image features and fed them to the decoder.
CNN is useful for recognizing small characters while preserving locality such as character positions, reducing the size of image features, and improving the computational efficiency and performance of the decoder.

The number of convolutional layers contributes to recognition performance, and many derivatives have been explored to increase it.
ResNet~\cite{Res15}, which consist of a large number of residual blocks of multiple convolutional layers with simple skip connections, has been commonly used.
In addition, ResNeXt~\cite{Res17} with group convolution and DenseNet~\cite{Dense17} with more complicated skip connections between all convolutional layers were proposed.

One of the weaknesses of CNN is its poor ability to recognize global context by focusing strongly on local features.
As a solution, a global context attention (GCA) block~\cite{GCA19} was proposed, defined by \eqref{GCA}.
\begin{equation}
\label{eq:GCA}
\bm{y}_{ij} = \bm{x}_{ij} + W_3 \max (0, \LayerNorm{} (W_2 \sum_i \sum_j \SoftMax{W_1 \bm{x}_{ij}}~\bm{x}_{ij})),
\end{equation}
where $i,j$ are the pixel indices, $\bm{x},\bm{y}$ are the input and output pixels, respectively.
$W_1,W_2,W_3$ are the weight matrices of three linear layers.
$\LayerNorm$ means layer normalization.
The softmax function is defined as follows.
\begin{equation}
\SoftMax{\bm{z}_{ij}} = \frac{\exp \bm{z}_{ij}}{\displaystyle \sum_m \sum_n \exp \bm{z}_{mn}},
\end{equation}
where $m,n$ are the pixel indices.
The GCA block should be placed between some residual (or dense) blocks.

\subsection{Decoder}

Transformer~\cite{Tr17} achieves superior performance in both language modeling and visual recognition tasks.
Compared to recurrent neural networks including long short-term memory~\cite{LSTM97}, a Transformer itself does not involve recursion, allowing parallel processing of sequential input and output data.
It should be noted that recurrent, sequential inference is performed unless the prediction length is fixed.
However, Transformer does not require recursion to recognize the context of the sequence, avoiding vanishing gradients and providing better performance.

The key idea of Transformer is called scaled dot product attention.
Let $X$ be a sequence of length $l_x$ and $d_x$ channels, and $Y$ be another input sequence.
For self attention, $X,Y$ are the same sequence, and the Transformer pays attention to other parts of $X$ in processing $X$.
For cross attention, $X$ and $Y$ are in different domains, and the Transformer pays attention to some parts from $Y$ in processing $X$.
These mechanisms allow Transformer to learn the context of sequential data and the relationship between visual and language domains.

The attention layer first generates a query $Q$, key $K$, and value $V$ from $X,Y$ as defined in \eqref{QKV}.
\begin{equation}
\label{eq:QKV}
\begin{array}{lll}
\bm{q}_i &= W_Q &\bm{x}_i, \\
\bm{k}_j &= W_K &\bm{y}_j, \\
\bm{v}_j &= W_V &\bm{y}_j, \\
\end{array}
\end{equation}
where $i,j$ are the sequence indices, $\bm{q},\bm{k},\bm{v},\bm{x},\bm{y}$ are the elements of $Q,K,V,X,Y$, respectively.
$W_Q, W_K, W_V$ are the projection matrices.

The output $Z$ of the attention layer is defined by \eqref{attn}.
\begin{equation}
\label{eq:attn}
Z_i = W_Z~\SoftMax{\frac{Q \trans{K}}{\sqrt{d_k}}} V,
\end{equation}
where $W_Z$ is the output projection matrix, and $d_k$ is the dimension of $\bm{k}$.
This is the mechanism of the scaled dot product attention~\cite{Tr17}.

In practice, the attention layer is divided into several groups, each of which pays attention independently and combines the outputs at the end.
This is called multi-head attention~\cite{Tr17}.
Through the above mechanism, Transformer can focus on specific values of $Y$ and incorporate them into the $X$ series.

\subsection{Local Attention}

Although Transformer has superior ability to recognize long sequences compared to recurrent neural networks, it is still known to perform poorly upon extremely long sequences.
Local attention (LA)~\cite{Tr20} is a technique designed to handle such long sequences in Transformer.

Let $M_{ij}$ be a mask to focus on the $j$th element from $i$th element of $X$.
The output of the local attention layer is defined by \eqref{mask}, involving causal masking to prevent leakage from subsequent elements.
\begin{equation}
\label{eq:mask}
Z_i = W_Z~\SoftMax{\frac{Q \trans{K}}{\sqrt{d_k}} + M_{ij}} V.
\end{equation}
The mask matrix $M$ is given by \eqref{local}.
\begin{equation}
\label{eq:local}
M_{ij} =
\begin{cases}
\hfill 0 & 0\leq i-j \leq w, \\
- \infty & \mathrm{otherwise},
\end{cases}
\end{equation}
where $i,j$ are the sequence indices, and $w$ is the width of the sliding window.

\subsection{Positional Encoding}

Transformer~\cite{Tr17} itself have poor ability to know the position of each element in the sequences, and the position information must be provided explicitly.
Instead of inputting simple position values, two approaches have been proposed, namely positional embedding~\cite{Pos17} and positional encoding~\cite{Tr17}.
In general, the latter works better on small training datasets.

The output $\bm{p}(n)$ of positional encoding for the index $n$ is defined by \eqref{pos}.
\begin{equation}
\label{eq:pos}
\bm{p}(n) =
\begin{pmatrix}
\vdots \\
\sin \frac{n}{10000^{\frac{2k}{d}}} \\[10pt]
\cos \frac{n}{10000^{\frac{2k}{d}}} \\
\vdots \\
\end{pmatrix},
\;
\text{where}~k \in \left[0, \frac{d}{2}\right).
\end{equation}
$\bm{p}(n)$ must be added directly to the feature vector $\bm{x}(n)$ with $d$ channels.

If the sequence $X$ has two-dimensional positions $(i,j)$, 2D positional encoding proposed by Zhao et al.~\cite{Pos21} may be a better choice.
It normalizes the horizontal and vertical coordinates to $[0,1]$, encodes each with \eqref{pos}, and then combines them to obtain a single vector.
The positional encoding of the $i,j$th pixel is given by \eqref{2d}.
\begin{equation}
\label{eq:2d}
\bm{p}_\mathrm{2D}(i,j) =
\begin{pmatrix}
\bm{p}\left(\frac{i}{H}\right) \\[10pt]
\bm{p}\left(\frac{j}{W}\right) \\
\end{pmatrix},
\end{equation}
where $H,W$ are the height and width for positional normalization.
In this paper, we omitted this normalization.

\subsection{Mutual Learning}

Ensemble learning is commonly used to improve machine learning generalization performance and fitting accuracy by averaging or complementing the outputs of multiple inference models.
However, it is computationally more expensive than single models due to the large number of parameters especially for deep learning approaches.

To achieve similar effects using only a single model, knowledge distillation~\cite{KD15} may be an alternative solution.
This is a technique which uses a large, complex neural network, i.e., an ensemble model, as a teacher and a small, simple model as a student to obtain higher performance than simply training a student model using the ground-truth data.

Mutual learning~\cite{Mut18} may be another solution.
Here, multiple student models are trained simultaneously to teach each other, without training a teacher model in advance.
In particular, each student model performs supervised learning using ground-truth data and minimizing Kullback--Leibler (KL) divergence~\cite{KL51} so that the distributions of each other's classification outputs match.

\subsection{Metrics\label{sec:TEDS}}

Zhong et al.~\cite{Zhong20} introduced a tree edit distance based similarity (TEDS) metric for performance evaluation of both table structure and cell content recognition.
After converting the recognition results and the ground truth into tree structures of HTML tags, the TEDS score is calculated according to \eqref{TEDS}.
\begin{equation}
\label{eq:TEDS}
\mathrm{TEDS}(T_a, T_b) = 1 - \frac{\EditDist{}(T_a, T_b)}{\max(|T_a|, |T_b|)},
\end{equation}
where $T_a$ and $T_b$ are the HTML trees, $\EditDist{}$ is the edit distance function, and $|T|$ is the number of nodes in $T$.

There are two versions of TEDS, namely structural TEDS and total TEDS.
The former is calculated for HTML trees excluding cell contents and represents the recognition performance for table structures only.
The latter is computed on complete HTML trees including cell contents and indicates the total recognition performance.

In addition, Zhong~\cite{Zhong20} classified the tables into two subsets, namely simple tables and complex tables.
The former are tables without cells which are merged vertically or horizontally, and the latter are the other tables.

\section{Proposal\label{sec:mutab}}

The proposal consists of a ResNet encoder and two local-attention Transformer decoders.
The two decoders infer table structure and cell contents, respectively.
An additional output layer estimates the cell bounding boxes.

\begin{figure}[tb]
\centering
\subfloat[CNN backbone.]{\includegraphics[width=.24\textwidth]{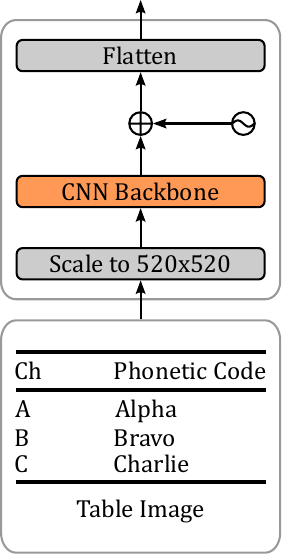}} \hspace{.05\textwidth}
\subfloat[HTML decoder.]{\includegraphics[width=.32\textwidth]{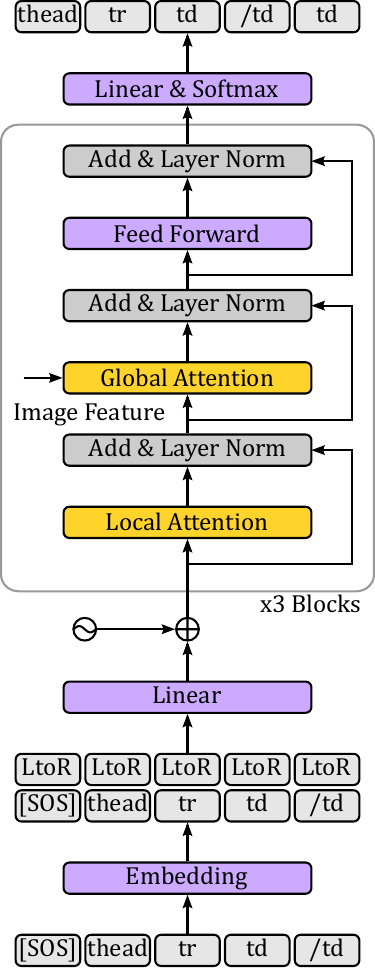}} \hspace{.05\textwidth}
\subfloat[Cell decoder.]{\includegraphics[width=.32\textwidth]{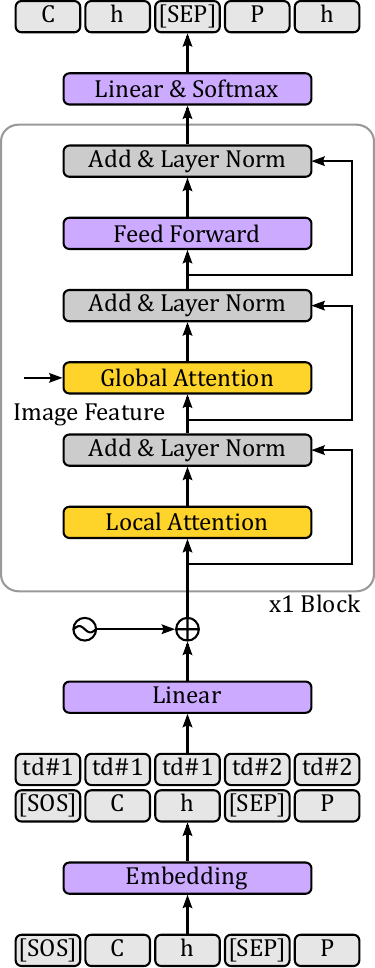}}
\caption{Proposed network architecture.\label{fig:arch}}
\end{figure}

The two main differences with previous studies~\cite{Nam23LA,Nam23GA} are 1) introduction of a multi-cell decoder, and 2) introduction of bidirectional mutual learning to the HTML decoder.
In addition, 2D positional encoding is employed.
We named the proposed method MuTabNet after mutual learning, multi-task learning, and the multi-cell decoder.
\figref{arch} shows the network architecture.

\subsection{Encoder}

The encoder consists of a CNN backbone and 2D positional encoding.
The CNN extracts image features of 65x65 pixels from an image of 520x520 pixels.
For the CNN, we adopted TableResNetExtra~\cite{VCG21} with 26 convolutional layers and three GCA blocks.
After 2D positional encoding, the image features are flattened into one-dimensional features with 512 channels for cross-attention at the decoders.

\subsection{HTML Decoder}

The HTML decoder consists of one embedding layer, three local attention blocks, and two output layers.
Each attention block accepts a table structure sequence through a self-attention layer in the block.
The attention block then incorporates image features into the table structure sequence through a cross-attention layer, and outputs the sequence through a feed-forward layer.
Several skip connections and layer normalizations are inserted within the block.
The output from the last attention block is converted into HTML tokens and cell bounding boxes by the two output layers.

During training, the decoder predicts left or right shifted HTML tokens from the input HTML tokens.
The shift direction is specified by an additional one-hot vector.
During inference, the decoder predicts the following token and iteratively expands the input sequence to obtain the complete HTML sequence.

In addition to HTML tokens, the decoder accepts some special tokens, namely \SOS{}, \EOS{}, and \PAD{}.
\SOS{} is a token that triggers sequential inference and is inserted at the beginning of the tokens.
\EOS{} is a token that stops inference and is inserted at the end of the tokens.
\PAD{} is inserted after \EOS{} to equalize the lengths of the tokens in the mini batch.

Following previous studies~\cite{Nam23LA,Nam23GA}, the HTML sequence was simply tokenized into HTML tags, except for the \TD{} tag representing the start of a cell.
A \TD{} tag is tokenized as `\TDL{}', `\colspan{}', `\rowspan{}', `\TDR{}' if it contains colspan or rowspan attributes.
Otherwise, the tag is simply tokenized as `\TD{}'.
It should be noted that FinTabNet~\cite{Zheng21} and PubTabNet~\cite{Zhong20} described in \secref{data} are publicly available with such tokenization applied.
We then merged the \TD{} and immediately following \TDE{} tokens into one token.

Furthermore, we assigned some special tokens to frequent sequence patterns.
If all header cells in the dataset had bold text, we removed the \B{} and completed it in post-processing.
These methods follow previous studies~\cite{Nam23LA,Nam23GA}.

\subsection{Cell Decoder}

The cell decoder consists of one embedding layer, one local attention block, and an output layer.
Following previous studies~\cite{Nam23LA,Nam23GA}, the embedding layer accepts cell characters one by one.
This is because cell contents typically consist of short sentences or unknown words or numbers, making it difficult to utilize pretrained language models.

The basic structure of a cell decoder is similar to that of an HTML decoder, with the following differences.
First, a special token \SEP{} is inserted between cell contents to trigger movement to the next cells.
Second, the cell decoder accepts a combination of cell contents and their corresponding HTML features extracted from the output of the HTML decoder.
These improvements allow the proposal to sequentially read the contents of multiple cells while referring to information from previous cells.

The previous study~\cite{Nam23LA} exploited local attention for the HTML decoder and global attention for the cell decoder.
This was because the cell decoder processed each cell independently, and the cell contents were short in general.
On the other hand, in our proposed multi-cell decoder, the sequence of cell contents tends to be long.
Consequently, we employed local attention.

\subsection{Bidirectional Mutual Learning}

We propose bidirectional mutual learning inspired by deep mutual learning~\cite{Mut18} to train the HTML decoder.
Here, two equivalent decoders are trained together to predict table structure in either a left-to-right (LtoR) or right-to-left (RtoL) direction.
To reduce model parameters, we implemented the mutual learning in a single decoder by combining an additional one-hot vector that determines the direction with the embedded HTML tokens.

Let $\LtoR{\bm{x}}$ and $\RtoL{\bm{x}}$ be the LtoR and RtoL sequences respectively, and let $p(\bm{x})$ be the ground-truth and $q(\bm{x})$ the predicted probabilities.
The loss $\LtoR{\mathcal{L}}$ for the LtoR decoder is defined by \eqref{LtoR}.
\begin{equation}
\label{eq:LtoR}
\LtoR{\mathcal{L}} =
- \frac{1}{N} \sum_{n=1}^N p(\LtoR{\bm{x}_n}) \log q(\LtoR{\bm{x}_n})
+ \frac{1}{N} \sum_{n=1}^N q(\RtoL{\bm{x}_n}) \log \frac{q(\RtoL{\bm{x}_n})}{q(\LtoR{\bm{x}_n})}.
\end{equation}

\section{Experiments}

To evaluate the effectiveness of the multi-cell decoder and bidirectional mutual learning, we conducted experiments on two public table datasets.

\subsection{Datasets\label{sec:data}}

We utilized two large datasets, FinTabNet~\cite{Zheng21} and PubTabNet~\cite{Zhong20}.
In addition, we used a subset named PubTabNet250~\cite{Nam23LA} for ablation studies.
\tabref{datasets} shows the statistics for the datasets.

\begin{table}[tb]
\centering
\caption{The statistics of the table image datasets.\label{table:datasets}}
\begin{tblr}{X[2]X[r]X[r]X[r]} \toprule
Dataset & Training & Validation & Evaluation \\ \midrule
FinTabNet    &  91,596 & 10,635 & 10,656 \\
PubTabNet    & 500,777 &  9,115 &  9,064 \\
PubTabNet250 & 114,111 &  2,161 &      - \\ \bottomrule
\end{tblr}
\end{table}

\subsubsection{FinTabNet}

is a large dataset of table images, including HTML labels and cell bounding boxes, extracted from the annual reports of S\&P 500 companies.
The dataset contains 112k tables and is divided into training set, validation set, and evaluation set.
It should be noted that the original FinTabNet confuses validation and evaluation sets.
Following previous studies~\cite{Nassar22,Zheng21}, we treated the \textit{validation} set containing 10,656 images as the evaluation set.

\subsubsection{PubTabNet}

is a dataset built by collecting scientific articles from the PubMed central open access subset, containing 568k tables and corresponding structure and cell content annotations and cell bounding boxes.
PubTabNet provides the training and validation sets, and the evaluation set was provided for the ICDAR competition~\cite{ICDAR21}.
We classified the tables into simple tables and complex tables as described in \secref{TEDS}.

\subsubsection{PubTabNet250}

Ly and Takasu~\cite{Nam23LA} extracted tables with 250 or more HTML tokens from PubTabNet and created a subset named PubTabNet250.
They also introduced subsets for tables containing at least 500, 600, and 700 tokens.
These subsets were utilized originally~\cite{Nam23LA} to demonstrate the effectiveness of the local attention mechanism.
We also utilized these subsets to conduct ablation studies in \secref{ablation}, approximately reducing training time from 179 hours to 45 hours per model.

\subsection{Implementation}

The proposed model was implemented in PyTorch using mmcv~\cite{MMCV}, mmdet~\cite{MMDET}, and mmocr~\cite{MMOCR} frameworks and trained on four NVIDIA V100 GPUs with batch size 8 in total.
We used Ranger~\cite{RANGER19} optimizer.
The learning rate was initialized to 0.001 for the first 25 epochs, and decreased to 0.0001 and 0.00001 for the next three and last two epochs, respectively.

Each tabular image was normalized and reduced to 520x520 pixels, padding the margins with zeros if necessary.
The cell bounding boxes were normalized to have a minimum value of 0 and a maximum value of 1.

HTML tokens and cell contents were converted to 512-dimensional embedded representations.
The four attention blocks in the HTML and cell decoders have the same 8-head, 512-channel architecture, and the sliding window size for local attention was set to 300 by default, following previous work~\cite{Nam23LA}.
The maximum lengths for table structure sequences and cell content sequences were set to 800 and 8000, respectively, including special tokens.
We employed greedy search for sequential prediction.

To ensure a fair comparison with the previous studies, we did not utilize data augmentation or ensemble learning techniques.
We also did not take advantage of early stopping.

\subsection{Experimental Results}

We compared the performance of the proposed model trained on FinTabNet and PubTabNet with the claimed performance of existing models.

\subsubsection{FinTabNet}

\begin{table}[tb]
\caption{Table recognition results on FinTabNet evaluation set.\label{table:fin}}
\begin{tblr}{colspec={XlX[c]X[c]}} \toprule
& & \SetCell[c=2]{c}TEDS (\%) \\ \cmidrule{3-4}
Model         &                 & Structure      & Total \\ \midrule
EDD           & \cite{Zhong20}  & 90.60          & - \\
GTE           & \cite{Zheng21}  & 87.14          & - \\
GTE (PT)      & \cite{Zheng21}  & 91.02          & - \\
TableFormer   & \cite{Nassar22} & 96.80          & - \\
VAST          & \cite{Huang23}  & 98.63          & \textbf{98.21} \\ \midrule
Ly et al.     & \cite{Nam23WS}  & 98.72          & 95.32 \\
Ly and Takasu & \cite{Nam23GA}  & 98.79          & - \\
Ly and Takasu & \cite{Nam23LA}  & 98.85          & 95.74 \\ \midrule
MuTabNet      &                 & \textbf{98.87} & 97.69 \\ \bottomrule
\end{tblr}
\end{table}

We evaluated the experimental results of structure recognition and total recognition using the TEDS metric.
\tabref{fin} compares the TEDS scores in the test set between the proposal and previous models.
The proposal outperforms the previous models with scores of 98.87\% and 97.69\%.
The inference time using the 4 GPUs was 3.78 hours.

The total TEDS score of the proposal was lower than the score of VAST~\cite{Huang23}, which could be explained by the fact that VAST exploits external OCR for cell content recognition.
In contrast, the structural TEDS score of VAST was lower than those of end-to-end approaches~\cite{Nam23WS,Nam23GA,Nam23LA}, including the proposal.

\subsubsection{PubTabNet}

\begin{table}[tb]
\centering
\caption{Table recognition results on PubTabNet validation set.\label{table:val}}
\begin{tblr}{colspec={XlX[c]X[c]X[c]}} \toprule
& & \SetCell[c=3]{c}TEDS (\%) \\ \cmidrule{3-5}
Model         &                 & Simple         & Complex        & Total \\ \midrule
EDD           & \cite{Zhong20}  & 91.20          & 85.40          & 88.30 \\
TabStruct-Net & \cite{Raja20}   &     -          &     -          & 90.10 \\
TableFormer   & \cite{Nassar22} & 95.40          & 90.10          & 93.60 \\
SEM           & \cite{Zhang22}  & 94.80          & 92.50          & 93.70 \\
LGPMA\&OCR    & \cite{Qiao21}   &     -          &     -          & 94.60 \\
VCGroup       & \cite{VCG21}    &     -          &     -          & 96.26 \\
VCGroup\&ME   & \cite{VCG21}    &     -          &     -          & 96.84 \\
VAST          & \cite{Huang23}  &     -          &     -          & 96.31 \\ \midrule
Ly et al.     & \cite{Nam23WS}  & 97.89          & 95.02          & 96.48 \\
Ly and Takasu & \cite{Nam23GA}  & 97.92          & 95.36          & 96.67 \\
Ly and Takasu & \cite{Nam23LA}  & 98.07          & 95.42          & 96.77 \\ \midrule
MuTabNet      &                 & \textbf{98.16} & \textbf{95.53} & \textbf{96.87} \\ \bottomrule
\end{tblr}
\end{table}

\begin{table}[tb]
\caption{Table recognition results on PubTabNet evaluation set.\label{table:test}}
\begin{tblr}{colspec={XlX[c]X[c]X[c]}} \toprule
& & \SetCell[c=3]{c}TEDS (\%) \\ \cmidrule{3-5}
Model         &                & Simple         & Complex        & Total \\ \midrule
LTIAYN        & \cite{ICDAR21} & 97.18          & 92.40          & 94.84 \\
anyone        & \cite{ICDAR21} & 96.95          & 93.43          & 95.23 \\
PaodingAI     & \cite{ICDAR21} & 97.35          & 93.79          & 95.61 \\
TAL           & \cite{ICDAR21} & 97.30          & 93.93          & 95.65 \\
DBJ           & \cite{ICDAR21} & 97.39          & 93.87          & 95.66 \\
YG            & \cite{ICDAR21} & 97.38          & 94.79          & 96.11 \\
XM            & \cite{Zhang22} & 97.60          & 94.89          & 96.27 \\
VCGroup       & \cite{VCG21}   & 97.90          & 94.68          & 96.32 \\
Davar-Lab-OCR & \cite{ICDAR21} & 97.88          & 94.78          & 96.36 \\ \midrule
Ly et al.     & \cite{Nam23WS} & 97.51          & 94.37          & 95.97 \\
Ly and Takasu & \cite{Nam23GA} & 97.60          & 94.68          & 96.17 \\
Ly and Takasu & \cite{Nam23LA} & 97.77          & 94.58          & 96.21 \\ \midrule
MuTabNet      &                & \textbf{98.01} & \textbf{94.98} & \textbf{96.53} \\ \bottomrule
\end{tblr}
\end{table}

We evaluated the experimental results of table recognition on the validation set using the TEDS metric.
\tabref {val} compares the scores between the proposal and previous methods.
The proposal outperforms all previous methods with scores of 98.16\%, 95.53\% and 96.87\% on simple tables, complex tables, and all tables, respectively.
The inference time using the 4 GPUs was 3.23 hours.

We also evaluated our proposal on the evaluation set.
\tabref{test} compares the scores of the proposal with the top 10 solutions of the ICDAR competition~\cite{ICDAR21}.
The high scores achieved on both sets indicate high generalization performance of the proposal.
The inference time was 3.13 hours.

The score of the proposal was higher than the score of VAST~\cite{Huang23}.
PubTabNet contains a large amount of training data, and the proposed model appears to be well trained for cell content recognition tasks.

It should be noted that VCGroup\&ME~\cite{VCG21} utilized additional annotation of bounding boxes of text lines within cell contents and ensemble learning of three models.
The proposed model outperforms all other non-end-to-end models which utilized additional annotation and ensemble learning even though our model did not utilize such techniques.

\subsection{Ablation Studies\label{sec:ablation}}

We conducted additional experiments for ablation studies using PubTabNet250 dataset for training and PubTabNet subsets for evaluation.

\subsubsection{Effectiveness of Multi-Cell Decoder and Mutual Learning}

We evaluated the effectiveness of the proposed methods, namely multi-cell (MC) decoder and bidirectional mutual learning (BML).
We trained two models on the training set and calculated the validation scores as displayed in \tabref{attn}.
We selected previous experimental results~\cite{Nam23LA} as baselines using exactly the same model architecture and dataset except for MC and BML.
LA in the table refers to local attention.

Since the previous study~\cite{Nam23LA} focused on performance for long tables, we also calculated TEDS scores for tables containing at least 500, 600, and 700 structure tokens.
The MC decoder outperforms the baselines at all table lengths, and BML further improves table recognition performance.

The effect of BML was unclear in the structural TEDS scores but evident in the total TEDS scores.
BML may still have improved the performance of implicit structure recognition and may have made an impact on cell content recognition, which requires precise content locations.

\subsubsection{Window Size of Cell Decoder}

Ly and Takasu~\cite{Nam23LA} has reported that a window size of 300 was optimal for the HTML decoder, whereas the cell decoder exploited global attention.
In this study, we determine the optimal window size for the MC decoder.
\tabref{window} shows the change in TEDS scores for the validation set as the window size varies from 100 to 500, while the window size for the HTML decoder was fixed at 300.

In general, a window size of 300 achieved the highest score, with the exception of tables containing more than 500 tokens, where a window size of 100 achieved the highest score.
Tables with many cells tend to have fewer characters per cell, and a shorter window may be sufficient.

It should be noted that we used the PubTabNet250 dataset for training, and the performance for tables with fewer structure tokens was lower than the scores in \tabref{val}.
We selected the window size of 300 as the optimal value for the entire PubTabNet dataset containing tables with fewer tokens from the perspective of generalization performance.

\begin{table}[tb]
\centering
\caption{Table recognition results with the proposed methods.\label{table:attn}}
\begin{tblr}{colspec={cccX[c]X[c]X[c]X[c]X[c]X[c]X[c]X[c]}} \toprule
& & & \SetCell[c=8]{c}TEDS (\%) \\ \cmidrule[l]{4-11}
\SetCell[c=3]{c}Methods & & & \SetCell[c=4]{c}Structure & & & & \SetCell[c=4]{c}Total \\ \cmidrule{1-3} \cmidrule[l]{4-7} \cmidrule[l]{8-11}
LA  & MC  & BML & 250+  & 500+  & 600+  & 700+  & 250+  & 500+  & 600+  & 700+  \\ \midrule
-   & -   & -   & -     & -     & -     & -     & 93.86 & 91.16 & 90.63 & 88.65 \\
\CM & -   & -   & -     & -     & -     & -     & 94.28 & 92.99 & 91.29 & 89.61 \\ \midrule
\CM & \CM & -   & 96.60 & 96.71 & 96.75 & 96.67 & 95.02 & 94.59 & 93.73 & 93.14 \\
\CM & \CM & \CM & 97.02 & 96.70 & 96.35 & 96.65 & 95.81 & 95.11 & 94.05 & 94.02 \\ \bottomrule
\end{tblr}
\end{table}

\begin{table}[tb]
\centering
\caption{Table recognition results with respect to window sizes.\label{table:window}}
\begin{tblr}{colspec={X[c]X[c]X[c]X[c]X[c]X[c]X[c]X[c]X[c]X[c]}} \toprule
& \SetCell[c=9]{c}TEDS (\%) \\ \cmidrule{2-10}
& \SetCell[c=4]{c}Structure & & & & \SetCell[c=5]{c}Total \\ \cmidrule{2-5} \cmidrule[l]{6-10}
Size & 250+  & 500+  & 600+  & 700+  & 0+    & 250+  & 500+  & 600+  & 700+  \\ \midrule
100  & 96.96 & 96.69 & 96.98 & 96.60 & 75.91 & 95.70 & 95.19 & 94.99 & 94.35 \\
200  & 96.79 & 96.53 & 96.30 & 95.83 & 75.79 & 95.46 & 94.66 & 93.80 & 92.69 \\
300  & 97.02 & 96.70 & 96.35 & 96.65 & 83.15 & 95.81 & 95.11 & 94.05 & 94.02 \\
400  & 96.83 & 96.85 & 96.48 & 96.51 & 82.58 & 95.40 & 95.08 & 94.00 & 93.50 \\
500  & 96.97 & 96.74 & 97.03 & 96.54 & 81.14 & 95.51 & 94.46 & 93.88 & 92.65 \\ \bottomrule
\end{tblr}
\end{table}

\section{Conclusion}

We improved an end-to-end table recognition model based upon Transformer to achieve performance comparable to state-of-the-art models using external OCR systems.
The proposed model consists of a ResNet encoder and two decoders for structure recognition and cell content recognition.
After the first decoder infers the structure tokens, the second decoder reads the text within each cell.

We proposed a multi-cell decoder for cell content recognition to exploit useful information from neighbor cells.
Furthermore, we proposed bidirectional mutual learning to force the model to pay attention to both previous and following cells.
Experimental results using two public datasets demonstrate the effectiveness of the proposed methods.

In future work, we will further consider multitasking models that include the task of recognizing the meaning of tables, which enables deep understanding of printed documents, including table contents, and provides high-quality scientific knowledge for LLMs and question-answering systems.

\bibliographystyle{splncs04}
\bibliography{icdar24}

\end{document}